# Exploring How EFL Students Talk To and Through AI to Develop Texts

David James Woo [a], Yangyang Yu [b, *], Yilin Huang [c], Deliang Wang [d], Kai Guo [e], and Chi Ho Yeung [c]

[a] Everwrite Limited, Hong Kong, China

[b] Shanghai Jiao Tong University, Shanghai, China

[c] The Education University of Hong Kong, Hong Kong, China

[d] The University of Hong Kong, Hong Kong, China

[e] The Chinese University of Hong Kong, Hong Kong, China

*** Corresponding author**

1. Email address: florayu0209@sjtu.edu.cn

2. Postal address: School of Foreign Languages, Shanghai Jiao Tong University, No. 800 Dongchuan Road, Minhang District, Shanghai, China, 200240

**Funding Statement**

The work by CHY is supported by the Research Grants Council of the Hong Kong Special Administrative Region, China (Projects No. GRF 18300623), and the Dean's Research Fund of the Faculty of Liberal Arts and Social Sciences, The Education University of Hong Kong, Hong Kong Special Administrative Region, China (Projects No. FLASS/DRF 04847).

**Acknowledgement**

The authors kindly thank Dr Hengky Susanto and Bohan Yang for their assistance.

## Declarations of interests

The authors report there are no competing interests to declare.

## Data availability statement

The data supporting this study's findings are available from the first author, David James Woo, upon reasonable request.

## Declaration of Generative AI and AI-assisted technologies in the writing process

During the preparation of this work, the authors used ChatGPT in order to improve readability and language. After using this tool, the authors reviewed and edited the content as needed and take full responsibility for the content of the publication.

# Exploring How EFL Students Talk To and Through AI to Develop Texts

**Abstract**

Generative artificial intelligence (AI) introduces new considerations for English as a Foreign Language (EFL) writing pedagogy. This study explores how students *talk to* and *through* AI by prompt engineering and negotiating authorship, respectively, and whether any patterns in the latter relate to students' writing performance. Using an exploratory mixed-methods design, we analyzed screen recordings of 44 Hong Kong secondary students completing a curricular writing task with AI chatbots. Content analysis identified 10 types of prompting strategies students employed, including questions, searches, and detailed instructions. From clustering these strategies, three distinct profiles of human-AI rhetorical load responsibility emerged: AI-dominant (52% of students), human-dominant (25%), and collaborative human-AI (14%). A MANOVA analysis indicated no significant multivariate effect of rhetorical load responsibility on three dimensions of students' writing performance: content, language, and organization. Students' prompting strategies and rhetorical load responsibility patterns have implications for their engagement and autonomy and EFL writing pedagogy.

## 3. Introduction

Generative artificial intelligence (AI) tools like ChatGPT have enhanced English as a Foreign Language (EFL) writing pedagogy, as research shows their use can improve EFL students' grammatical accuracy, content organization, and task completion (Marzuki et al., 2023; Oktarin et al., 2024). At the same time, these tools' increasing capabilities, for example, to act autonomously without human intervention (Xi et al., 2025), are creating more complex models of human-AI collaboration that go beyond traditional machine-in-the-loop writing (Clark et al., 2018), where AI plays a supporting role and the human writer maintains full agency for the final written product. On the one hand, it is possible that EFL students can successfully navigate this complex human-AI collaborative process (Hutson, 2025) in a way that is not detrimental to their agency and learning. On the other hand, research suggests that even the mere use of large language models (LLMs) during the writing process creates cognitive debt and reduced human agency compared to writing without digital tools and to writing with search engines (Kosmyna et al., 2025).

To help EFL students navigate the complex human-AI collaborative process, this paper explores how students are positioning themselves relative to two dimensions when writing with AI. First, how students *talk to AI* through prompt engineering, that is, the metacognitive and metalinguistic process of crafting instructions to generate desired output from AI. Second, how students *talk through AI* or how they negotiate authorship and agency with AI while writing, which determines the extent to which a final composition could be considered human-authored or wholly synthetic (Hau, 2025). By exploring how Hong Kong EFL secondary school students' prompting strategies reflect different patterns of human-AI agency, we aim to understand if any patterns relate to students' writing performance. By identifying these relationships, we might develop more effective prompt engineering instruction and approaches to student agency that enhance learning outcomes.

## 4. Literature Review

### *4.1. From Process Writing to Machine-in-the-loop Writing*

EFL writing instruction typically has approached composition writing as a process with iterative stages of planning, drafting, and revising (Hyland, 2019). Negotiating these stages has required complex cognitive activity (Flower & Hayes, 1981). The introduction of generative AI tools into process writing adds iterative, cognitive stages, what has been coined machine-in-the-loop writing (Clark et al., 2018): an EFL student writes prompts or instructions to the AI tool; evaluates the AI tool's output; integrates the output into the composition as necessary; and writes the composition independent of AI support. On the one hand, the addition of AI tools could enhance students' metacognitive skills as students consider why they prompt AI tools (Woo, Guo, et al., 2024) and how to best use the tools to support the writing process (Fathi & Rahimi, 2024). On the other hand, students' over-reliance on AI tools could increase their cognitive debt (Kosmyna et al., 2025) and cognitive load (Woo, Wang, et al., 2024) that inhibit learning outcomes. In the next sections, we propose that how students talk to AI while writing is a metacognitive practice that has implications for student agency, cognition, and writing performance.

### *4.2. 'Talking to AI' as Prompt Engineering Metacognitive Practice*

'Talking to AI' refers to prompt engineering or crafting appropriate instructions so that AI generates the user's desired output. It requires metacognitive and metalinguistic skills, that is, conscious reasoning and reflection on communication goals and language use. Appropriate instructions can be crafted from natural language processing (NLP) scaffolds that enhance AI's understanding. For instance, a user could greet and chat with an AI as if it were a human, asking it a question, or issuing a command (Ouyang et al., 2022). Additionally, users can provide examples of task performance, although AI may perform some tasks proficiently without any examples (Brown et al., 2020). From a study of Hong Kong EFL students' prompt engineering strategies, Woo, Susanto, et al. (2025) identified that students could treat AI as if it were a Google search engine, and that students could utilize prompts autonomously generated by AI. Importantly, how EFL students 'talk to AI' or apply different NLP scaffolds to compose prompt engineering strategies can reflect not only their metacognitive and metalinguistic skills, but also the use case. For EFL writing tasks, students' prompt engineering strategies may reflect the extent to which students use AI-generated text in their writing (Woo, Susanto, et al., 2025).

### *4.3. 'Talking through AI' as Distributing Agency and Rhetorical Load*

'Talking through AI' refers to negotiating authorship and voice in completing a human-AI composition. In an EFL writing context, AI-generated text in student compositions (Woo, Susanto, et al., 2024, 2025) has been a means to determine final authorship. However, we know little about how EFL students negotiate authorship while writing. Humans and AI could assume different production roles while writing, even on a moment by moment basis (Goffman, 1981). Significantly, Knowles (2024) proposes Rhetorical Load Theory to conceptualize how cognitive and rhetorical responsibilities are distributed between collaborators. The theory ultimately distinguishes between an ethical 'machine-in-the-loop' writing where humans retain primary rhetorical responsibility with AI providing support and 'human-in-the-loop' writing where AI carries most rhetorical load and humans merely monitor. Whether humans or AI carry the rhetorical load can inform whether a composition should be considered human-authored or synthetic.

Recent research has demonstrated that how students talk through AI may influence their writing performance. Woo, Susanto, et al. (2024) analyzed EFL secondary students' compositions written with AI, finding that the number of AI-generated words and the number of human words in the final compositions significantly influenced their writing performance. Another study by Woo, Susanto, et al. (2025) revealed that whether AI use could lead to higher-quality writing may depend on how students interacted with AI-generated text, which often varied with EFL students' baseline writing competence. Additionally, some research has connected real-time student-AI interaction while writing with writing outcomes. Kim et al. (2025) revealed that master students with higher AI literacy tended to adopt a more collaborative approach to writing with AI and achieve better essay task scores. Nguyen et al. (2024) suggested that doctoral students collaborating with AI in an iterative and interactive way were more likely to excel in academic writing. Both studies detailed dynamic postgraduate student-AI interaction patterns from screen recordings that captured students' writing behaviors, but neither of them concerned younger EFL learners or examined authorship negotiation during the prompting stage. Thus, it remains unexplored how younger EFL students' talking to AI may translate to their talking through AI from the perspective of distributing agency and rhetorical load responsibility, and whether these factors lead to a difference in these students' writing performance.

The present study conceptualizes EFL students' talking to AI and talking through AI as interrelated phenomena. Specifically, we explore how Hong Kong EFL secondary students' prompts grant agency and rhetorical load responsibility to AI, with each prompt shaping whether the machine is in the loop or the EFL student is in the loop while writing. At present, we know little about younger EFL students' prompts that empower AI with agency and rhetorical load responsibility, AI's specific agentic roles and rhetorical load responsibility in the writing process, and the extent to which rhetorical load responsibility impacts students' writing performance. Because of these, the following are our research questions (RQs):

- ***RQ1***: What types of NLP scaffolds do Hong Kong EFL students use in their prompts to AI chatbots?
- ***RQ2***: What are the prompting strategies that Hong Kong EFL students employ with AI chatbots?
- ***RQ3***: What type of agentic roles do EFL students grant AI through their prompting strategies?

- ***RQ4***: To what extent do students and AI bear rhetorical load responsibility?
- ***RQ5***: What are the relationships between rhetorical load responsibility and EFL students' writing performance, if any?

## 5. Methodology

### *5.1. Research Context and Participants*

To sample a wide range of EFL learners, this study was conducted in five Hong Kong secondary schools, which enroll students at different levels of overall academic achievement including EFL achievement. Forty-four students participated. They were informed of the study and their rights and could withdraw from the study at any time.

Students attended a two-hour workshop at their school. The workshop's title was 'How to attempt a writing task with ChatGPT support' (see Appendix A for the workshop learning design). The first author led the workshop. Students were introduced to an explicit EFL writing approach (genre-based or process-based) selected by their English teacher. They received prompt engineering instruction to support that approach to writing. Next, they received a feature article writing task (see Figure 1) chosen by their English teacher from three possible questions taken from a Hong Kong Diploma of Secondary Education (HKDSE) examination writing paper that all Hong Kong mainstream school students must take in Form 6 (US Grade Level 12, Ages 17-18). Their compositions could not exceed 500 words.

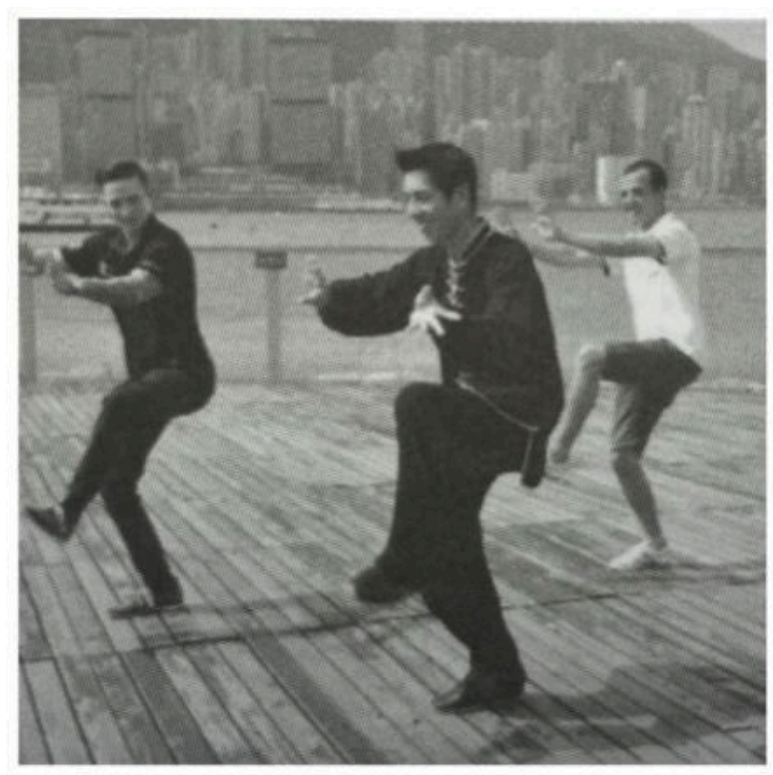

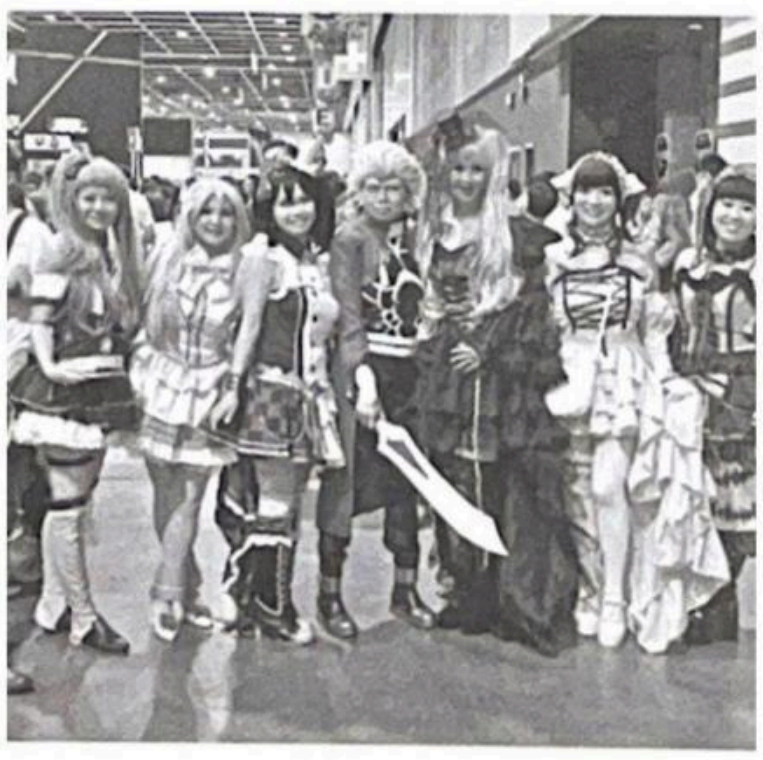

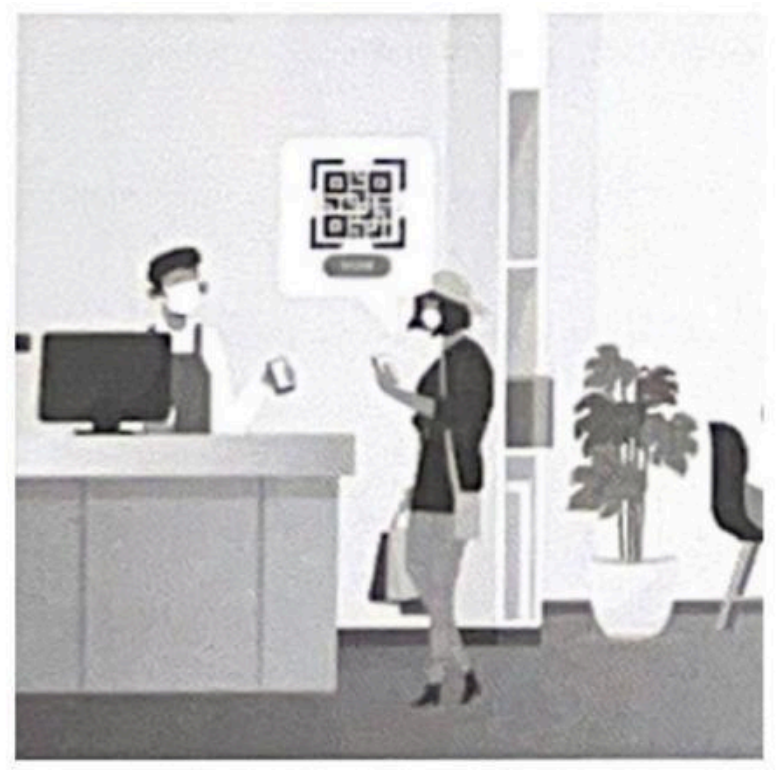

**Question 1**

**Learning English through Sports Communication**

While tai chi is a popular activity in Hong Kong, it is less known in some parts of the world.

Write an article for *International Travel* magazine introducing the benefits of tai chi to tourists.

**Question 2**

**Learning English through Popular Culture**

Anime Expo, Hong Kong's biggest anime, manga and video game exhibition, was held at the Hong Kong Convention and Exhibition Centre last weekend. As a school reporter, you attended the event and interviewed some people dressed in cosplay.

Write an article for your school magazine.

**Question 3**

**Learning English through Workplace Communication**

You work for *Restaurant Business* magazine. You interviewed a restaurant owner about his/her experiences of running a business during the pandemic.

Write a feature article for the magazine.

**Figure 1.** Three possible questions for a student's feature article writing task

Students were instructed to use their own words and words from at least one AI chatbot in their composition. Students could use as many or as few of their own words or AI words as necessary. Although students were introduced to the POE app and its collection of state-of-the-art chatbots in the workshop (see Figure 2), students could use other chatbot applications and as many chatbots as necessary. Students attempted the task according to their school-supplied hardware, either on an iPad, laptop, or desktop. Students wrote their texts on Google Docs and shared these with the first author.

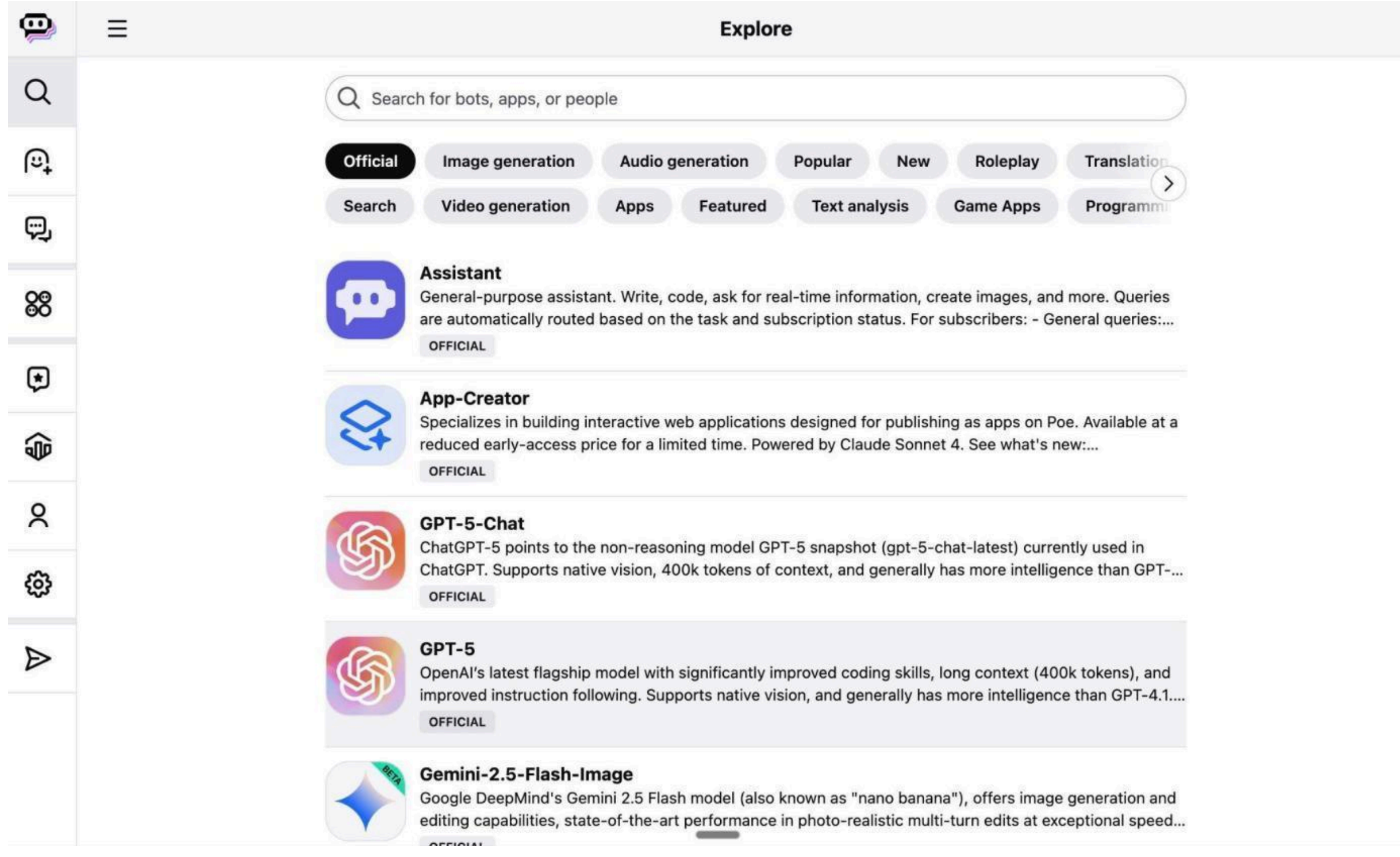

**Figure 2.** The POE app interface

### *5.2. Data Collection*

This study adopts an exploratory sequential mixed methods design, summarized in Table 1, where qualitative methods inform quantitative methods (Creswell & Plano Clark, 2018). First, we collected 44 screen recordings from students' hardware as they completed the writing task. In addition, we collected 27 student texts that did not exceed the 500 word limit. In the next section, we elaborate the initial analysis of screen recordings to answer RQ1, which informs subsequent RQs' data sources and analysis.

**Table 1.** Exploratory Sequential Mixed Methods Design Overview

| Framework | RQs | Data source | Data analysis approach |
|---|---|---|---|
| Talk to AI | RQ1: What types of NLP scaffolds do Hong Kong EFL students use in their | Screen recordings of student prompts | Content analysis of student prompts<br>- Deductive-inductive coding |
| | prompts to AI chatbots? | (n=384 total prompts) | - Three-round coding process with 100% agreement |
| | RQ2: What are the prompting strategies that Hong Kong EFL students employ with AI chatbots? | NLP scaffold frequency data | Frequency analysis<br>- Code dominance identification<br>- Threshold analysis (>0.3) |
| Talk through AI | RQ3: What type of agentic roles do EFL students grant AI through their prompting strategies? | Code dominance patterns | Qualitative content and thematic analysis<br>- Code clustering into AI role categories<br>- Assignment of role codes to students |
| | RQ4: To what extent do students and AI bear rhetorical load responsibility? | AI role codes + analytic memos | Constant comparison method<br>- Merging into rhetorical load responsibility profiles<br>- Iterative comparison of prompts and memos |
| Performance integration | RQ5: What are the relationships between rhetorical load responsibility and EFL students' writing performance, if any? | Student texts + rhetorical load profiles | Multi-scorer assessment + MANOVA<br>- Dual-scorer evaluation using Hong Kong curricular marking scheme<br>- Blinded scoring for content, language, organization |

## *5.3. Data Analysis*

### *5.3.1. For Addressing RQ1*

To answer what types of NLP scaffolds Hong Kong EFL students employ with AI chatbots, we conducted a content analysis (Joffe & Yardley, 2004) on the screen recordings, focusing on the prompting stage. For each screen recording, the third author assigned a student identifier (i.e., school, form-level, and number information) and noted the total runtime. She transcribed each prompt in the recording verbatim. For each prompt, she assigned a sequential number, and documented the timestamp and which chatbot received the prompt. She also wrote analytic memos (e.g., student reactions, editing behaviors, and use of other AI tools and reference materials). The first author subsequently triangulated and validated these memos by reviewing the original screen recordings, correcting or annotating the memos where necessary to ensure accuracy. This data was organized in a spreadsheet to facilitate systematic coding.

We developed a coding scheme to capture the NLP scaffolds in students' prompts. To ensure coding validity and reliability, the first and third authors jointly developed a codebook (Saldana, 2012). We aimed to develop exclusive codes, that is, to assign one NLP scaffold code per prompt. We employed a combined deductive-inductive coding approach. We started with deductive codes adopted from an LLM instruction-following taxonomy (Ouyang et al., 2022), prompt content types (Woo, Guo, et al., 2025), and zero shot, one shot, and few shot examples (Brown et al., 2020). Additionally, we developed inductive codes that emerged from prompts in the screen recordings and that were not covered in the existing scheme.

For the first round of coding, the third author applied deductive codes and developed inductive codes. For independent verification, the first author randomly selected 25% of the prompts (n=95) to code, agreeing on 63% of the prompt coding (n=60). The first and third authors reviewed their codes and revised code definitions in the codebook. Then the first author re-coded in a second round. To check for reliability, the first author selected another 25% of prompts to code, this time achieving 100% agreement on code application. We present descriptive analyses of the codes.

### *5.3.2. For Addressing RQ2*

To compose students' prompting strategies, we quantified code frequencies to identify dominant patterns: we composed the frequency counts of the prompt codes per student, calculated proportions of each code relative to a student's total code count, and identified a student's dominant codes, if any, using a researcher-designed threshold (i.e. >0.3) (see Supplemental Material for the code dominance table).

### *5.3.3. For Addressing RQ3*

To identify AI's agentic roles, we performed a qualitative content analysis (Vaismoradi & Snelgrove, 2019), aiming to iteratively identify a small set of categories linking prompting strategies to AI agency. First, we reviewed the code dominance results and developed categories by grouping dominant codes, and assigned these category codes to students. We added these category codes and descriptors to the codebook. We present descriptive analyses of the AI agentic role codes.

### *5.3.4. For Addressing RQ4*

To address the extent that students and AI bear rhetorical load responsibility, we aimed to merge the category codes into a smaller set of thematic codes that reflect distinct rhetorical load responsibility profiles and assign them to students. To elaborate each theme, the first and third authors led the constant comparison analysis (Glaser & Strauss, 1967), iteratively comparing students' prompts and analytic memos against these themes to describe them, and open coding analytic memos for common student responses for a theme (see Supplemental Material for analytic memo codes). Theme definitions and student assignments were refined through discussion between the authors. We present descriptive analyses of the rhetorical load responsibility thematic codes with representative prompts and common student responses to AI-generated output.

#### *5.3.5. For Addressing RQ5*

To explore any relationship between rhetorical load responsibility and writing performance, the first author and students' English teachers scored students' compositions using the HKDSE writing marking guidelines (see Appendix B). The guidelines assess students' writing performance by evaluating each text's content, language, and organization on a 0-7 scale. Before scoring, the third author excluded texts that did not follow the task instruction (e.g., exceeding the word limit), and anonymized the remaining qualified texts (n=27). The two scorers independently awarded scores for each qualified text without knowing the extent of AI-generated words in the text. Their scores were averaged to reach final scores for each dimension of the text.

We then conducted Multivariate Analysis of Variance (MANOVA) on participants who had both text scores and classified rhetorical load responsibility profiles, to examine whether rhetorical load responsibility was associated with writing performance in content, language, and organization. Statistical analyses and visualizations were performed in R (version 4.5.0).

## 6. Findings

### *6.1. RQ1: Types of NLP Scaffolds*

We observed all students prompting chatbots in the POE application. ChatGPT (n=206) and Assistant (n=70) received the most prompts. Students also prompted derivative ChatGPTS (e.g., ChataiGPT, ChatGPTfree, n=49), Google chatbots (n=39), and Llama-2-70b (n=30).

We developed 10 possible NLP scaffolds per Table 2. For our deductive codes, we found students most frequently asked questions (code Q, n=76) followed by using chatbots to search for information (code SE, n=71). Students least frequently used chatbots for autocompleting a prompt without natural language instruction (code AC, n=1). We developed four inductive codes, observing students writing prompts comprising natural language instruction with details or context (code DNLI, n=81), requesting the chatbot to rewrite existing text (code RE, n=62), writing prompts comprising natural language instruction with details or context and a question (code Q+DNLI, n=10), and using the chatbot as an auxiliary tool, for instance, to count words or to translate (code AT, n=5). DNLI prompts were written by the greatest number of unique students (n=32), followed by SE (n=27) and Q (n=26) prompts. We observed AT (n=3) and AC (n=1) prompts were written by the least number of unique students. Five prompts were unclassified.

**Table 2.** Coding Scheme and Frequencies of Natural Language Processing (NLP) Scaffolds in Student Prompts

| Code type | Prompt code | Description | Example prompts | No. of instances | No. of students |
|---|---|---|---|---|---|
| Deductive | AC | Auto-complete: a one-shot prompt of an input data exemplar without natural language instruction and without a question. The intent is for the AI to learn to generate output for the input data exemplar by autocomplete, that is, extension. | hence Pivoting to Survival: Creative Strategies and Innovations (approx. 150 words) Detail the innovative measures Anderson took to adapt his business model to the new reality. Discuss the implementation of online ordering systems, delivery services, and curbside pickup to maintain revenue streams. Highlight any unique initiatives Anderson introduced, such as virtual cooking classes or meal kits, to engage with customers and generate additional income. (Student copied a part of AI generated before and added "hence" before first word) | 1 | 1 |
| Deductive | GRE | Greeting: a prompt that acknowledges the chatbot as if it were a human but that does not appear to facilitate completion of the writing task. | thank u for ur response | 18 | 7 |
| Deductive | NLI | Natural language instruction: explicit instruction in natural language with an imperative verb and ZERO shot, that is, without an input detail. NLI to rewrite should be coded to RE. | Write a openingabout the above feature article | 14 | 9 |
| Deductive | Q | Question: a prompt formulated with either a direct or indirect question, that is, a statement beginning with a question word. Typically an information question. Not longer than two clauses. Questions to rewrite should be coded to RE. | what is the formate of a magazine | 76 | 26 |
| Deductive | SE | Search engine: a prompt for information, grounded in the real world. Factual, not formulated as a question. Does not need an imperative verb. Can include requests for meanings and examples, summaries, backgrounds, and histories but exclude | benefit of Tai Chi<br><br>Summaries the benefit of Tai Chi<br><br>History of Tai Chi | 71 | 27 |
| | | NLP tasks like summarize. | | | |
| Deductive | TMM | It not only means "tell me more" automatically generated by AI, but also contains other questions automatically generated and recommended by AI. | Are there any well-known health organizations or sports associations that conduct surveys on Tai Chi's popularity. | 24 | 7 |
| Inductive | AT | Auxiliary tools: Use AI as an auxiliary tool, such as making AI number, translation, word count, etc. | can you help me to count the number of wor<br><br>all words in article | 5 | 3 |
| Inductive | DNLI | Natural language instruction with details. A typical example is the writing task prompt. NLI to rewrite should be coded to RE. | I mean the ideas shouldn't be specfic to one topic, the ideas of what to write for each paragragh shoule be just what to write for the feature article | 81 | 32 |
| Inductive | Q + DNLI | A question with generative intent, and natural language instruction or context or details for at least two clauses long. | can you generate a model text of an article? i am a reporter from my school who went to the anime exhibition and interviewed cosplayers. | 10 | 9 |
| Inductive | RE | Rewrite: ask AI to rewrite the answers. Can imply a reference to previous prompt(s) and output(s). Can include details for rewriting. Can include additions and merges to existing writing. Can include a question to rewrite. | in like 50 words<br><br>delete the part B of the introduction and the part of the interviews with cosplayers<br><br>can you add some parts as if you interviewed some people dressed in cosplay? | 62 | 23 |
| Inductive | * | Unable to log prompt or invalid prompt | (The video provided by the student did not capture the prompt)<br><br>(The length of the prompt was beyond the length of the context window and the video did not capture the entire prompt.) | 5 | 0 |

### *6.2. RQ2: Prompting Strategies*

We identified 40 students who were dominant in at least one NLP scaffold. Figure 3 shows these students' distribution across their dominant NLP scaffold(s). Twenty-five students showed dominance in only one scaffold, ten students in two scaffolds and five in three. The most dominant scaffold was DNLI (n=18), followed by RE (n=14) and SE (n=11). No students showed dominance in AC and AT.

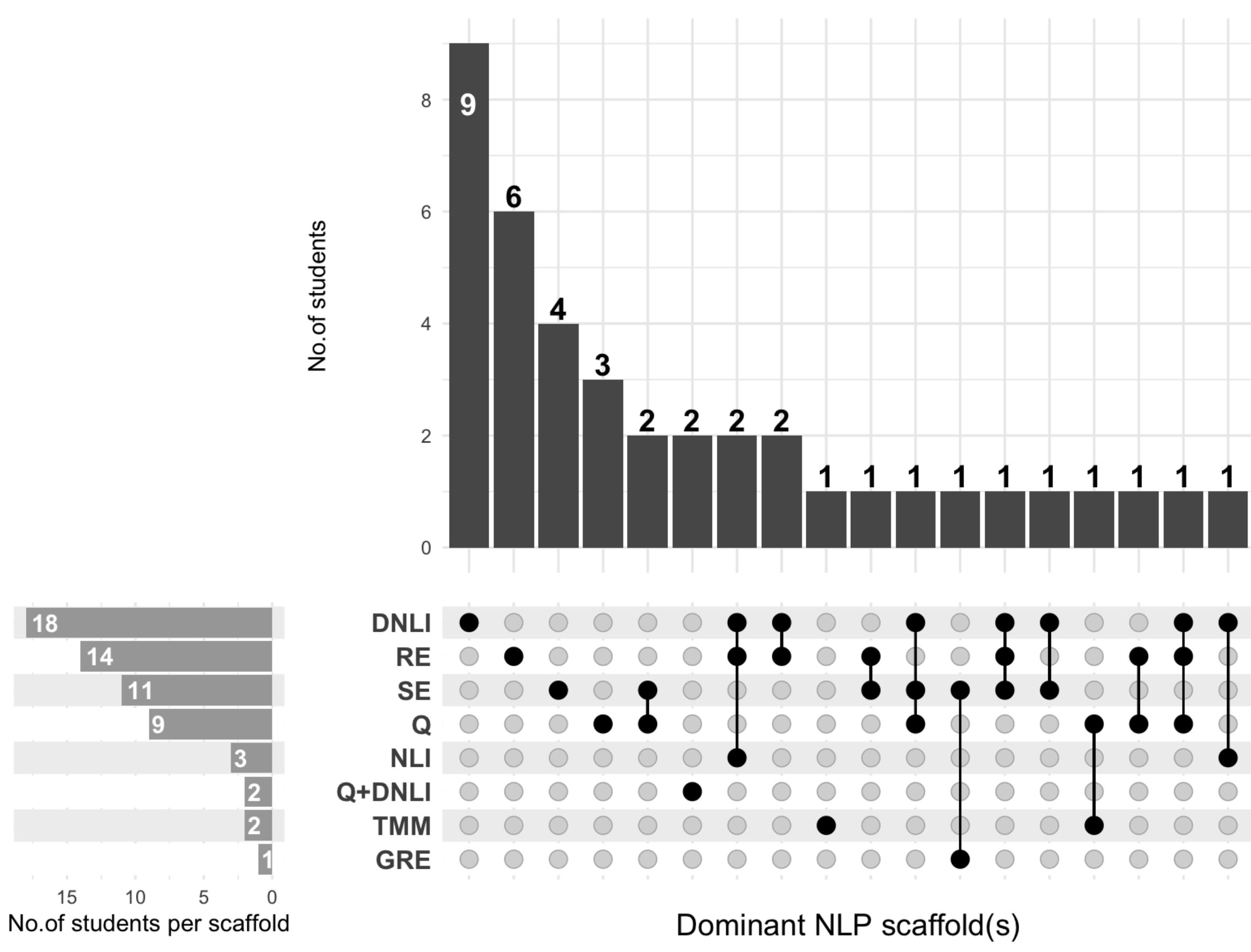

**Figure 3.** Distribution of 40 students across their dominant NLP scaffolds

*Note.* Left bars represent the number of students per dominant NLP scaffold. Top bars show the number of students in each intersection of dominant scaffold(s). Filled dots below the top bars indicate which scaffold(s) are included in each intersection.

### *6.3. RQ3: Types of AI Agentic Roles*

We developed four categories which describe the AI chatbot's role while writing. Figure 4 presents these categories with the distribution of students. *Article Generator* (code

AG from prompt codes DNLI, NLI, Q+DNLI, RE) referred to using the AI chatbot as a text generator and was assigned to 28 students (64%); *Search Engine* (code SG from prompt codes SE, Q) referred to using the AI chatbot as a search engine such as Google and was assigned to 17 students (39%); *Auxiliary Tools* (code AT from prompt codes AC, TMM, AT) referred to using the chatbot as an auxiliary tool and was assigned to two students (5%); and *Partner* (code P from prompt code GRE) referred to talking with the AI chatbot as if it were a friend and was assigned to one student (2%). Four students (9%) showed no dominant prompt code and were unclassified. Notably, eight students applied AI for multiple roles, such as combining Article Generator and Search Engine (n=6), Search Engine and Auxiliary Tools (n=1), or Partner and Search Engine (n=1).

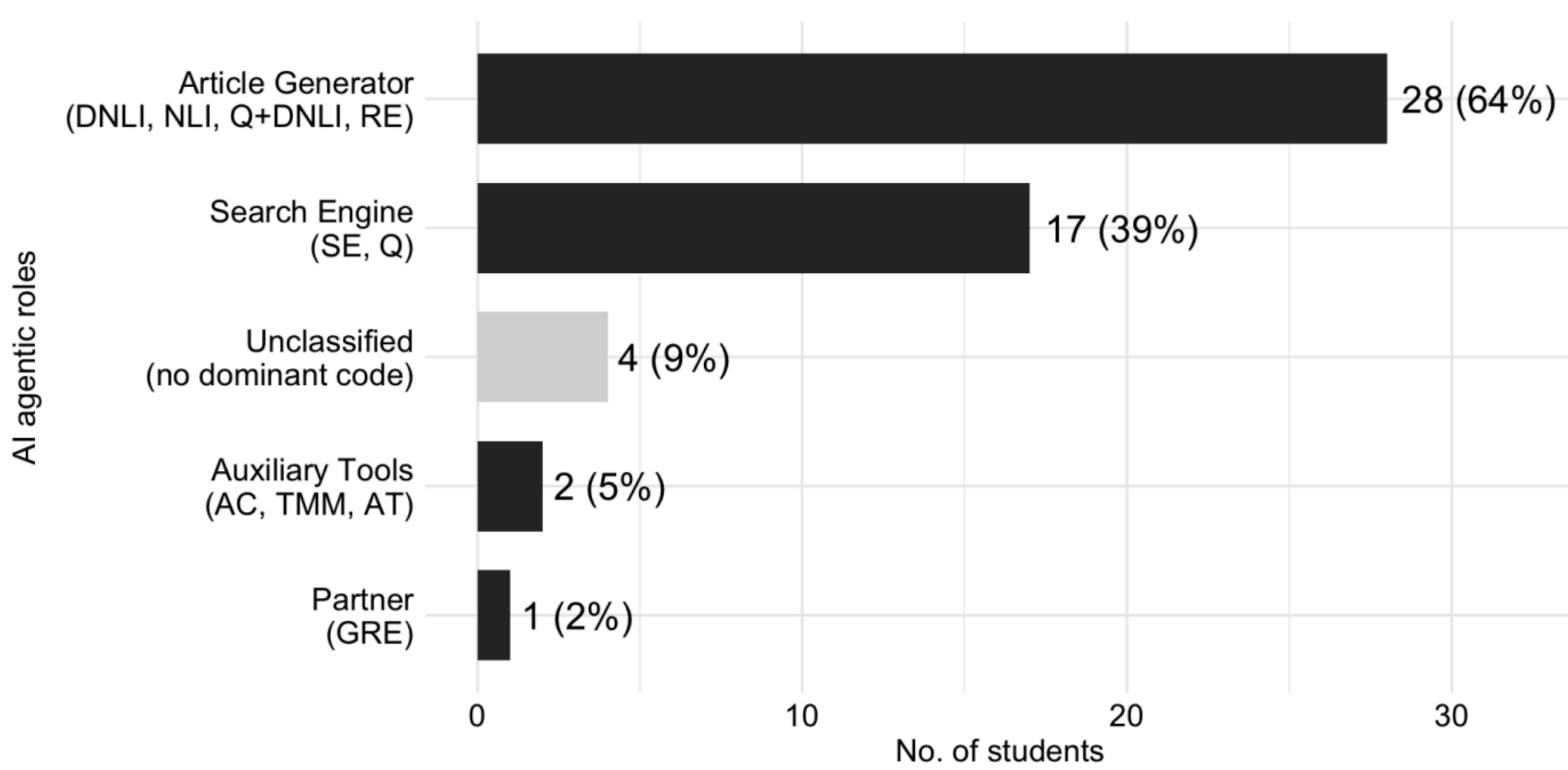

**Figure 4.** Distribution of 44 students across different types of AI agentic roles

### *6.4. RQ4: Rhetorical Load Responsibility Profiles*

We merged the AI role codes into three themes that describe a spectrum of human-AI rhetorical load responsibility. We describe the themes below.

***Theme 1: AI.*** This theme (from role code AG) indicated that the AI primarily bore the rhetorical load responsibility and was assigned to 23 students (52%). Per the AG role code, students prompted a chatbot with direct commands like "Write a writing plan for a feature article…Not more than 500 words," followed by "Write a feature article for the magazine... Not more than 500 words" and then "hence write about adapting to changing circumstances (approx. 100 words)" (hpccss_4B05); and "Generate a writing plan for an article…," followed by "generate an introduction for this writing plan in about 80 words…" and then "following the writing plan, generate the third part…in about 150 words" (spslt_4A20). Furthermore, analytical memos showed students' lack of engagement with AI-generated output as the majority of students in this cohort (n=14) copied and pasted the entirety of an AI-generated output into their composition at least once, and four students did so several times. These students appeared reliant on AI for idea generation and content creation, with little critical engagement or revision.

***Theme 2: Human.*** This theme (from AI role codes AT, SE, P) indicated that the student primarily bore the rhetorical load responsibility and was assigned to 11 students (25%). Per the Q and SE role codes, students asked AI for information including "what is Tai Chi," "how popular is Tai Chi HK," and "why people all over the world only knows a little about Tai Chi" (hpccss_4B15), or searched for information such as "Give some examples on the effects of tai chi, and explain," "explain how tai chi effects ones health," and "list 5 benefits of tai chi" (hpccss_4B21), respectively. Analytical memos showed students' critical engagement. For example, most students in the cohort (n=8) appeared to show dissatisfaction with AI-generated output at least once and seven students did several times. Dissatisfaction could come in the form of rephrasing prompts or testing prompts on several chatbots. For instance, one student (hpccss_4B17) submitted a prompt ("benefits of tai chi to tourist") to ChatGPT and, immediately after, attempted to access Meta's Llama and Google's Palm LLMs, subsequently returning to ChatGPT and writing a more elaborate prompt ("plot me a plan of a letter to the editor about the benefits of tai chi to tourist"). Another student (hpccss_4B01) submitted a prompt ("Give me some aspects of benefits and elaborate that Tai Chi can bring to the tourism industry") to Assistant, skimmed its output, and immediately submitted the same prompt to ChatGPT. The student skimmed the new output, highlighted and right-clicked an output section, and selected "dislike" from the right-click menu. In these ways, it appeared that students intended to establish their voice through ideas and content.

***Theme 3: Human-AI.*** This theme (from a combination of role code AG and AT, SE or P) indicated that the human and AI shared rhetorical load responsibility with neither dominant and was assigned to six students (14%). Students prompted AI to play more than one agentic role, which we observed in consecutive prompts when a student would use one prompt scaffold and then another. For example, one student prompted "generate information about tai chi such as its origin. List them in point form" followed by "how to write a opening sentence in an article to grab readers attention. Give me 2 examples" (hpccss_4D29). In other cases, students prompted "Write an article…Write no more than 500 words" followed by "why do people love cosplay" (spslt4_D02), and "awe-inspiring synonym" followed by "rewrite the second last paragraph" (spslt4_D20). The analytical memos showed a pair of students (n=2) copied and pasted the entirety of an AI-generated output, but three other students (n=3) showed critical engagement with AI-generated output. With this type of agency, AI and students tended to give and take for idea generation and content creation, with neither dominating.

### *6.5. RQ5: Rhetorical Load Responsibility and Writing Performance*

Twenty-seven out of the 44 participants submitted qualified texts that received human-rated scores. After excluding three with unclassified rhetorical load responsibility profiles, 24 participants' text scores and classified profiles were included in MANOVA. The distribution of profiles among these participants was: 11 in the 'AI' group, 9 in the 'Human' group, and 4 in the 'Human-AI' group. The independent variable was the profile type, while the dependent variables were content, language and organization scores.

The MANOVA indicated no significant multivariate effect of rhetorical load responsibility on three dimensions of writing performance, Pillai's Trace = .18, $F(6,40) = 0.66$ $p = .68$. Follow-up univariate ANOVAs likewise revealed no significant group differences for content, $F(2, 21) = 0.09$, $p = .91$, language, $F(2, 21) = 0.01$, $p = .99$, organization, $F(2, 21) = 0.10$, $p = .91$. All effect sizes were negligible (partial $\eta^2 < .01$). Figure 5 further illustrates substantial overlap among the three groups in the scores, confirming that rhetorical load responsibility did not meaningfully affect students' writing performance.

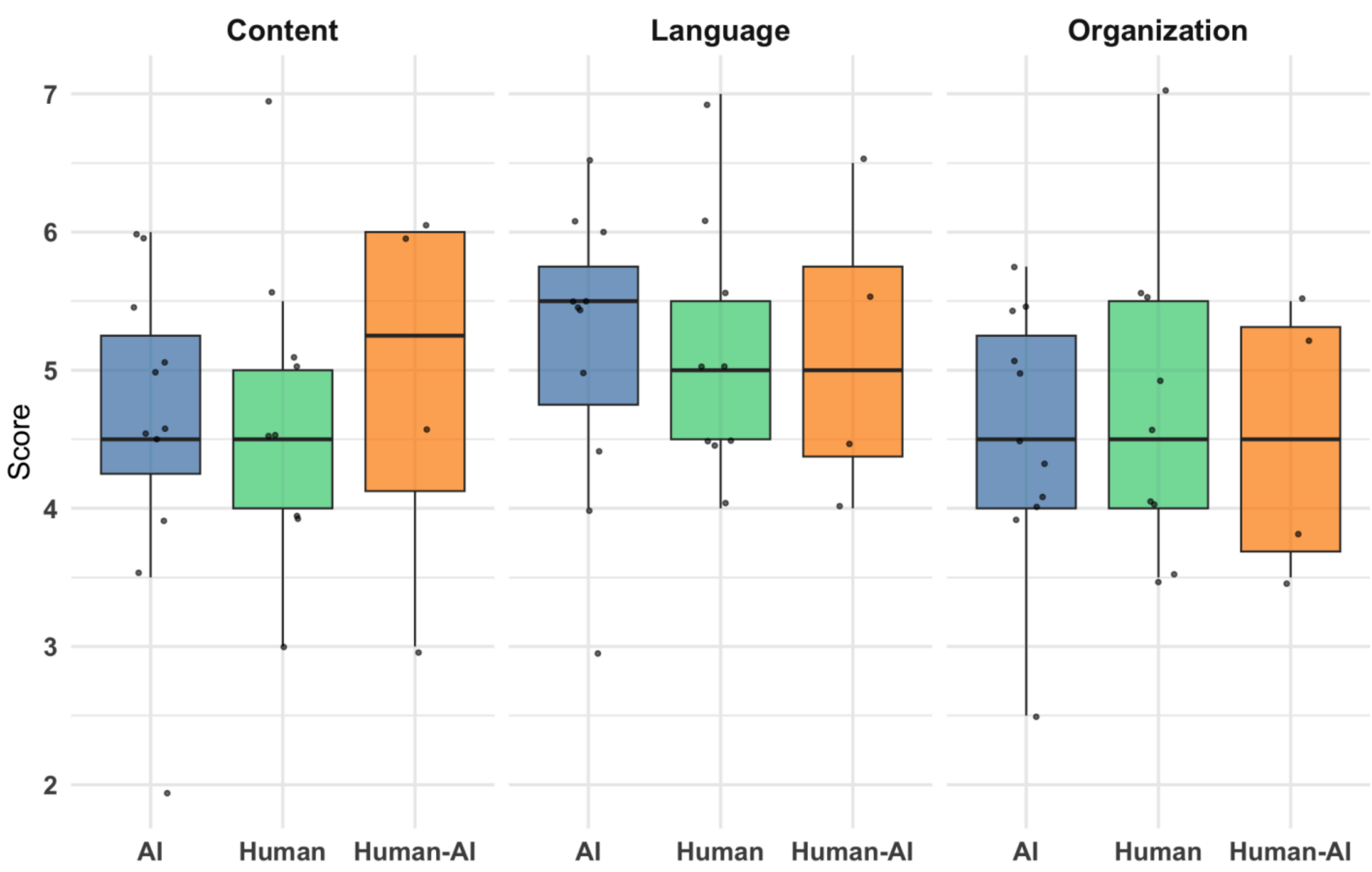

**Figure 5.** Students' writing performance by rhetorical load responsibility profile (AI, n = 11; Human, n = 9; Human-AI, n = 4; Total, N = 24)

## 7. Discussion

This study explored how Hong Kong EFL secondary students engage with AI chatbots to complete a feature article writing task, examining how they *talk to AI* through prompts and *talk through AI* by negotiating authorship. The findings indicate that students engaged in significant metalinguistic and metacognitive practice while *talking to AI*. Prevalent NLP scaffolds including detailed instructions (DNLI) and rewrites (RE) indicate students' awareness of how formulating language shapes outputs and their critical evaluation and refinement of responses, respectively. Questions (Q) and searches (SE) show students' actively seeking knowledge. These frequently employed scaffolds by EFL secondary students resonate findings that higher education students interacted with GenAI for various purposes throughout the writing process such as idea generation, linguistic support, and information searching (Hwang et al., 2025; Nguyen et al., 2024). That 15 students employed multiple dominant strategies further suggests students' capacity for flexible and sophisticated interaction with AI at the prompt engineering stage. Thus, it appears that how students are talking to AI benefits their learning.

However, how students talk to AI does not necessarily translate to much student agency and authorship when *talking through AI*. First, AI agentic roles show concerning patterns and diminished student agency. The Article Generator role (64% of students), comprising DNLI and RE prompting strategies, was dominant, intending students' delegation of authorship to AI. Although the Search Engine (39% of students), Auxiliary Tools (5% of students) and Partner (2% of students) may better enhance student agency and show a range of possible agentic roles, these agentic roles were rare compared to AI as an Article Generator. The diminished student agency may coincide with "metacognitive laziness" (Fan et al., 2024), where students offload metacognitive effort to AI by positioning it more as Article Generator. Without proper instruction and guidance on AI use, students might gradually reduce their cognitive engagement with the writing process and eventually lose effective self-regulation in writing development.

The three rhetorical load profiles reveal a prevailing pattern of students' over-reliance on AI to carry rhetorical load responsibility at the cost of learning. The majority of students adopting an AI-dominant profile (52% of students) may stem from a combination of factors. First, students may have received insufficient training that limits their awareness of prompting strategies. Alternatively, the time-sensitive task along with negotiating the writing process with AI may have increased students' cognitive load (Woo, Wang, et al., 2024) and caused students to seek an expedient path to task completion. In that way, the AI-dominant profile facilitates a paradoxical human-in-the-loop writing process where students skillfully control prompts when talking to AI, with the aim of ceding rhetorical responsibility when talking through AI, presumably to punctually complete the composition and to relieve cognitive load. Importantly, the AI-dominant profile may be detrimental to learning, leading to students' cognitive debt as short-term task completion takes precedence over long-term development of writing skills like idea generation and revision. In contrast, the other two profiles appear more representative of writing with a machine in the loop and suggest more ethical and productive models of human-AI collaboration for learning. Both human-dominant (25% of students) and human-AI collaborative (14% of students) profiles demonstrate that students can write with AI to supplement, rather than supplant, their own authorship and cognition over the writing process. However, these students constituted only a minority.

While Knowles (2024) advocates machine-in-the-loop writing as the ideal collaborative model for students writing development, the MANOVA analysis indicated no significant multivariate effect of rhetorical load responsibility on three dimensions of students' writing performance: content, language, and organization. The null finding suggests that rhetorical load responsibility, whether AI-dominant, human-dominant, or human-AI collaborative, neither degrades nor enhances students' writing performance on this particular task. This could be a result of a traditional assessment rubric lacking sensitivity to capture quality differences emerging from different rhetorical load profiles, just as the same rubric appears insensitive to capturing quality differences from the pasting of AI-generated text (Woo, Susanto, et al., 2025). Significantly, assessments of EFL student writing that are agnostic towards human-AI agency may ultimately hinder students' development of writing skills independent of AI use. Alternatively, differences in students' individual traits, such as their general literacy skills, AI literacy, or motivation (Kim et al., 2025; Jin et al., 2025), may exert stronger influences on their writing outcomes. These individual factors may have interacted with, or even overshadowed, any subtle effects of rhetorical load responsibility within our relatively limited sample size, resulting in comparable writing scores across the different groups. Although the present MANOVA contrasted with prior research that proved the impact of student-AI interaction patterns on writing performance (Nguyen et al., 2024; Kim et al., 2025), the analysis of prompting strategies and the inclusion of rhetorical load responsibility as a potential factor may still extend our understanding of machine-in-the-loop writing in a more nuanced and theoretically grounded way.

The three rhetorical load responsibility profiles illustrate different ways of talking through AI, not only expanding the range of support available to EFL students but also exposing them to the challenges of managing agency and authorship in the writing process. Therefore, educators play an important role in guiding students through this complex human-AI collaborative writing process. First, educators should not assume that sophisticated prompting alone fosters learning, nor should they expect rhetorical load responsibility to predict writing task performance. Practically, given the prevalence of the AI-dominant profile, EFL educators should aim for students to balance their agency with AI. That can begin with teaching the writing process without generative AI, developing students' agency, and then introducing generative AI in a gradual, scaffolded manner. Concrete strategies include: (1) aligning prompt engineering education with writing stage; for instance, using Q and SE for idea generation, DNLI for planning, and RE for revision rather than full-text drafting; (2) implementing a protocol like “Accept, Adapt, Reject” to guide students in critically analyzing AI output, facilitating students’ rhetorical responsibility; and (3) incorporating checkpoints where students reflect on who contributed which parts of a draft and why. Educators may design assessment rubrics that reward or penalize different rhetorical load profiles during the writing process. These strategies may promote student engagement in the writing process and mitigate students’ cognitive debt from wholesale copying.

Although the findings have implications for EFL pedagogy and should lead educators to rethink their prompt engineering education and the agentic roles that students assign to AI, several limitations constrain their generalizability. The study’s context is specific to Hong Kong secondary students, and concerns a short-duration task. Our analysis drew on screen recordings and memos, which captured observable behavior but not students’ private reasoning. The exploration into human-AI agency and writing performance was particularly limited to 24 cases. In addition, the study did not incorporate individual learner traits, which could have influenced writing outcomes and interacted with rhetorical responsibility profiles.

## 8. Conclusion

This study explored how EFL students talk to and through AI, showing that students’ prompt engineering strategies correspond to different models of rhetorical responsibility. It revealed a paradox: EFL students are developing sophisticated ways of talking to AI, but without intervention they often talk through AI in ways that outsource rhetorical responsibility. Crucially, EFL students will need explicit guidance from educators to address this gap so that AI use ultimately supports rather than compromises language learning.

**Appendix A.** Workshop learning design

| Title | How to attempt a writing task with ChatGPT support |
|---|---|
| Contact time | 2 hours |
| Purpose | To develop Hong Kong students' and teachers' competence to use ChatGPT for English writing enhancement |
| Intended learning outcomes (LT) // Essential understandings | 1. I can understand genre / process and its approach to writing.<br>2. I can understand ChatGPT and identify its tasks<br>3. I can understand prompts and identify their categories<br>4. I can write prompts for different writing stages<br>5. I can independently develop a text with the support of ChatGPT |
| Learning activities (ILOs) (minutes) | 1. Pre-workshop questionnaire (5 minutes)<br>2. Introduction to writing approach (B1. Genre-based / B2. Process-based) (10 minutes)<br>3. Introduction to AI, chatbots and ChatGPT (5 minutes)<br>4. Model prompt types with examples (25 minutes)<br>5. Guided practice applying prompts to writing stages for an HKDSE task (25 minutes)<br>6. Introduction to contest and setting up (10 minutes)<br>7. Independent practice on HKDSE writing task (30 minutes)<br>8. Wrapping up and post-workshop questionnaire (10 minutes) |
| Materials (written language) | 1. Generative AI tools on POE app on iPads<br>2. Google Docs<br>3. Shared Google Drive folder:<br>a. Contest website (English language)<br>b. Marking scheme (English language)<br>c. Pre- and post-workshop questionnaires (English and Chinese languages)<br>d. Workshop slidedeck (English language)<br>e. Worksheets (English language)<br>4. iPads / desktops<br>5. Poll Everywhere (English language) |
| Instructional language | English |

**Appendix B.** HKDSE English Language Paper Two (Writing) Marking Scheme.

| Marks | Content (C) | Language (L) | Organization (O) |
|---|---|---|---|
| **7** | · Content entirely fulfills the requirements of the question<br>· Totally relevant<br>· All ideas are well developed/supported<br>· Creativity and imagination are shown when appropriate<br>· Shows a high awareness of audience | · Very wide range of accurate sentence structures, with a good grasp of more complex structures<br>· Grammar accurate with only very minor slips<br>· Vocabulary well-chosen and often used appropriately to express subtleties of meaning<br>· Spelling and punctuation are almost entirely correct<br>· Register, tone and style are entirely appropriate to the genre and text-type | · Text is organized extremely effectively, with logical development of ideas<br>· Cohesion in most parts of the text is very clear<br>· Cohesive ties throughout the text are sophisticated<br>· Overall structure is coherent, extremely sophisticated and entirely appropriate to the genre and text-type |
| **6** | · Content fulfills the requirements of the question<br>· Almost totally relevant<br>· Most ideas are well developed/supported<br>· Creativity and imagination are shown when appropriate<br>· Shows general awareness of audience | · Wide range of accurate sentence structures with a good grasp of simple and complex sentences<br>· Grammar mainly accurate with occasional common errors that do not affect overall clarity<br>· Vocabulary is wide, with many examples of more sophisticated lexis<br>· Spelling and punctuation are mostly correct<br>· Register, tone and style are appropriate to the genre and text-type | · Text is organized effectively, with logical development of ideas<br>· Cohesion in most parts of the text is clear<br>· Strong cohesive ties throughout the text<br>· Overall structure is coherent, sophisticated and appropriate to the genre and text-type |
| **5** | · Content addresses the requirements of the question adequately<br>· Mostly relevant | · A range of accurate sentence structures with some attempts to use more complex sentences | · Text is mostly organized effectively, with logical development of ideas<br>· Cohesion in most parts of the text is clear<br>· Sound cohesive ties throughout the text |
| | · Some ideas are well developed/supported<br>· Creativity and imagination are shown in most parts when appropriate<br>· Shows some awareness of audience | · Grammatical errors occur in more complex structures but overall clarity not affected<br>· Vocabulary is moderately wide and used appropriately<br>· Spelling and punctuation are sufficiently accurate to convey meaning<br>· Register, tone and style are mostly appropriate to the genre and text-type | · Overall structure is coherent and appropriate to the genre and text-type |
| **4** | · Content just satisfies the requirements of the question<br>· Relevant ideas but may show some gaps or redundant information<br>· Some ideas but not well developed<br>· Some evidence of creativity and imagination<br>· Shows occasional awareness of audience | · Simple sentences are generally accurately constructed.<br>· Occasional attempts are made to use more complex sentences.<br>· Structures used tend to be repetitive in nature<br>· Grammatical errors sometimes affect meaning<br>· Common vocabulary is generally appropriate<br>· Most common words are spelt correctly, with basic punctuation being accurate<br>· There is some evidence of register, tone and style appropriate to the genre and text-type | · Parts of the text have clearly defined topics<br>· Cohesion in some parts of the text is clear<br>· Some cohesive ties in some parts of the text<br>· Overall structure is mostly coherent and appropriate to the genre and text-type |
| **3** | · Content partially satisfies the requirements of the question<br>· Some relevant ideas but there are gaps in candidates' understanding of the topic | · Short simple sentences are generally accurate.<br>· Only scattered attempts at longer, more complex sentences<br>· Grammatical errors often affect meaning<br>· Simple vocabulary is appropriate | · Parts of the text are generally defined<br>· Some simple cohesive ties used in some parts of the text but cohesion is sometimes fuzzy<br>· A limited range of cohesive devices are used appropriately |
| | · Ideas not developed, with possible repetition<br>· Does not orient reader effectively to the topic | · Spelling of common words is correct, with basic punctuation mostly accurate | |
| **2** | · Content shows very limited attempts to fulfil the requirements of the question<br>· Intermittently relevant<br>· Some ideas but few are developed<br>· Ideas may include misconception of the task or some inaccurate information<br>· Very limited awareness of audience | · Some short simple sentences accurately structured<br>· Grammatical errors frequently obscure meaning<br>· Very simple vocabulary of limited range often based on the prompt(s)<br>· A few words are spelt correctly with basic punctuation being occasionally accurate | · Parts of the text reflect some attempts to organize topics<br>· Some use of cohesive devices to link ideas |
| **1** | · Content inadequate and heavily based on the task prompt(s)<br>· A few ideas but none developed<br>· Some points/ ideas are copied from the task prompt or the reading texts<br>· Almost total lack of awareness of audience | · Multiple errors in sentence structures, spelling and/or word usage, which make understanding impossible | · Some attempt to organize the text<br>· Very limited use of cohesive devices to link ideas |
| **0** | · Totally inadequate<br>· Totally irrelevant or memorized<br>· All ideas are copied from the task prompt or the reading texts<br>· No awareness of audience | · Not enough language to assess | · Mainly disconnected words, short note-like phrases or incomplete sentences<br>· Cohesive devices almost entirely absent |